\documentclass{IEEEoj-data}
\usepackage{cite}
\usepackage{amsmath,amssymb,amsfonts}
\usepackage{algorithmic}
\usepackage{graphicx,color}
\usepackage{textcomp}
\usepackage{hyperref}
\usepackage{balance}
\usepackage{booktabs}
\usepackage{multirow}
\usepackage{makecell}

\usepackage[acronym,toc]{glossaries}

\def\BibTeX{{\rm B\kern-.05em{\sc i\kern-.025em b}\kern-.08em
    T\kern-.1667em\lower.7ex\hbox{E}\kern-.125emX}}
\AtBeginDocument{\definecolor{ojcolor}{cmyk}{0.93,0.59,0.15,0.02}}
\glsdisablehyper

\newacronym{lidar}{LiDAR}{Light Detection and Ranging}
\newacronym{bev}{BEV}{Bird's Eye View}
\newacronym{radar}{RADAR}{Radio Detection And Ranging}
\newacronym{sota}{SOTA}{State-of-the-art}
\newacronym{iou}{IoU}{Intersection-over-Union}
\newacronym{miou}{mIoU}{Mean Intersection-over-Union}
\newacronym{cptv}{CTPV}{ Cylindrical Triperspective View}
\newacronym{cit}{CIT}{Cross-modal Interaction Transform}
\newacronym{dff}{DDF}{Dual Dynamic Fusion}
\newacronym{mos}{MOS}{moving object segmentation}
\newacronym{vim}{ViM}{Vision Mamba}
\newacronym{vit}{ViT}{Vision Transformer}
\newacronym{rgb}{RGB}{Red, Green, Blue}
\newacronym{bce}{BCE}{binary cross entropy}
\newacronym{ssim}{SSIM}{mean Structural Similarity Index}
\newacronym{fps}{FPS}{frames per second}
\glsaddall
\begin{document}

\title{\textcolor{black}{Descriptor:} \textcolor{ieeedata}{\textit{ Occluded nuScenes: A Multi-Sensor Dataset for Evaluating Perception Robustness in Automated Driving}} }

\author{SANJAY~KUMAR\authorrefmark{1,2,3}, (Graduate Student Member, IEEE), Tim~Brophy\authorrefmark{1,2,3} (Member, IEEE), Reenu~Mohandas\authorrefmark{1,2,3} (Member, IEEE), EOIN MARTINO GRUA\authorrefmark{1,2}, GANESH~SISTU\authorrefmark{1,4}, VALENTINA~DONZELLA\authorrefmark{5} (Senior Member, IEEE), CIARÁN~EISING\authorrefmark{1,2,3} (Senior Member, IEEE)}
\affil{Department of Electronic and Computer Engineering, University of Limerick, Limerick V94 T9PX, Ireland}
\affil{Data Driven Computer Engineering Research Centre, University of Limerick, Limerick V94 T9PX, Ireland}
\affil{Lero, The Irish Software Research Centre, University of Limerick, Limerick V94 NYD3, Ireland}
\affil{Valeo Vision Systems, Dunmore Road, Tuam, Co. Galway H54 Y276, Ireland}
\affil{Queen Mary University of London, Mile End Road, London E1 4NS, United Kingdom}
\corresp{CORRESPONDING AUTHOR: SANJAY KUMAR (e-mail: kumar.sanjay@ul.ie).}
\authornote{This work was supported in part by Science Foundation Ireland under Grant 13/RC/2094\_P2 and co-funded under the European Regional Development Fund
through the Southern and Eastern Regional Operational Programme to Lero - the Science Foundation Ireland Research Centre for Software, and in part by Valeo
Vision Systems.}
\markboth{Occluded nuScenes: A Multi-Sensor Dataset for Evaluating Perception Robustness in Automated Driving}{KUMAR \textit{et al.}}

\begin{abstract}
Robust perception in automated driving requires reliable performance under adverse conditions, where sensors may be affected by partial failures or environmental occlusions. Although existing autonomous driving datasets inherently contain sensor noise and environmental variability, very few enable controlled, parameterised, and reproducible degradations across multiple sensing modalities. This gap limits the ability to systematically evaluate how perception and fusion architectures perform under well-defined adverse conditions.
To address this limitation, we introduce the Occluded nuScenes Dataset, a novel extension of the widely used nuScenes benchmark. For the camera modality, we release both the full and mini versions with four types of occlusions, two adapted from public implementations and two newly designed. For radar and LiDAR, we provide parameterised occlusion scripts that implement three types of degradations each, enabling flexible and repeatable generation of occluded data. This resource supports consistent, reproducible evaluation of perception models under partial sensor failures and environmental interference. By releasing the first multi-sensor occlusion dataset with controlled and reproducible degradations, we aim to advance research on robust sensor fusion, resilience analysis, and safety-critical perception in automated driving.\\

 {\textcolor{ieeedata}{\abstractheadfont\bfseries{IEEE SOCIETY/COUNCIL}}}     IEEE Intelligent Transportation Systems Society (ITSS)\\  
 \\
 {\textcolor{ieeedata}{\abstractheadfont\bfseries{DATA DOI/PID}} \href{https://doi.org/10.21227/fd64-0p49}{10.21227/fd64-0p49}\\
  }
 {\textcolor{ieeedata}{\abstractheadfont\bfseries{DATA TYPE/LOCATION}}}  Images (.png); University of Limerick, Ireland

\end{abstract}


\maketitle

\section*{BACKGROUND} 

Automated driving systems depend on multi-sensor perception pipelines that integrate information from cameras, LiDAR, and radar to ensure an accurate and reliable understanding of the environment \cite{yan2024benchmarking}, \cite{gupta2025enhancing}. For use in safety-critical applications, these systems must remain robust under various challenging conditions, including adverse weather, sensor faults, and occlusions caused by objects in the driving scene. While recent fusion methods such as BEVFusion \cite{liu2022bevfusion}, TransFusion \cite{bai2022transfusion}, and BEVCar \cite{schramm2024bevcar} have demonstrated strong performance under clean conditions, their resilience to degraded sensor data remains underexplored.

Existing robustness studies and benchmarks have primarily focused on camera degradation \cite{wang2025msafusion}, \cite{zhang2024enhancing}, \cite{teresa2025deep}, \cite{xie2023robobev}, with limited consideration of radar and LiDAR. Datasets such as KITTI \cite{geiger2013vision}, Waymo Open Dataset \cite{sun2020scalability}, Argoverse \cite{chang2019argoverse}, and nuScenes \cite{caesar2020nuscenes} provide rich multi-modal recordings, but none provide controlled and parameterised occlusions across all three primary sensor modalities. As a result, the research community lacks a standardised and reproducible dataset to study how perception and fusion models behave when sensors are partially occluded or degraded.

To address this gap, we introduce the first dataset that applies synthetic occlusions across camera, radar, and LiDAR modalities within the widely used nuScenes \cite{caesar2020nuscenes} dataset. This resource enables consistent evaluation of perception models under partial sensor failure and occluded conditions, fostering the development of robust and resilient automated driving systems.

\section*{COLLECTION METHODS AND DESIGN} 
The occluded dataset is derived from the publicly available nuScenes dataset \cite{caesar2020nuscenes}, which provides synchronised multi-modal recordings collected in complex urban environments using a full-scale automated driving sensor suite. It includes six cameras, five Radars, a 32-channel LiDAR, and detailed 3D annotations and tracking. We selected nuScenes over other autonomous driving datasets such as KITTI \cite{geiger2013vision} and Waymo \cite{sun2020scalability} because it uniquely offers 360° calibrated multi-sensor coverage, including Radar, as well as a wide range of urban scenes captured under varying weather and lighting conditions. In contrast, KITTI lacks radar and is limited to front-facing sensors recorded under clear conditions, while Waymo also lacks radar and exhibits less environmental and lighting diversity. Although nuScenes provides comprehensive and well-calibrated sensor coverage, real-world deployments often experience partial fields of view and non-overlapping camera–LiDAR perspectives, which motivated our focus on simulating occlusion and partial sensor degradation.

To replicate real-world perception challenges such as lens contamination, adverse weather, and partial sensor failures, we developed a synthetic occlusion generation pipeline that introduces parameterised degradations across all three modalities. For cameras, we release preprocessed occluded images in both the full and mini versions of nuScenes, encompassing four types of visual occlusions. For radar and LiDAR, we provide parameterised scripts implementing three distinct degradation modes per modality, allowing users to generate occluded data at different severity levels in a flexible and reproducible manner. This design ensures full compatibility with existing nuScenes annotations and perception pipelines while offering researchers a controlled, reproducible framework for benchmarking perception robustness under degraded sensing conditions.

\subsection{Camera Occlusion}
We introduce four types of synthetic occlusions applied to the nuScenes camera images. Two of these (dirt and water-blur) are adapted from existing public repositories \cite{10900335} and parameterised to three severity levels, while two (WoodScape soiling patterns and scratches) are newly developed as part of this work. These were selected as they represent the common real-world camera occlusions caused by lens contamination and surface wear, with scratches introducing a novel distortion and WoodScape patterns replicating fog-like effects \cite{bijelic2020seeing}. All occlusions are applied to the raw camera streams, and the resulting occluded images are released for both the full and mini versions of nuScenes.

\subsubsection{Dirt Simulation}

\begin{figure*}[t]
\centering
\includegraphics[width=0.8\textwidth]{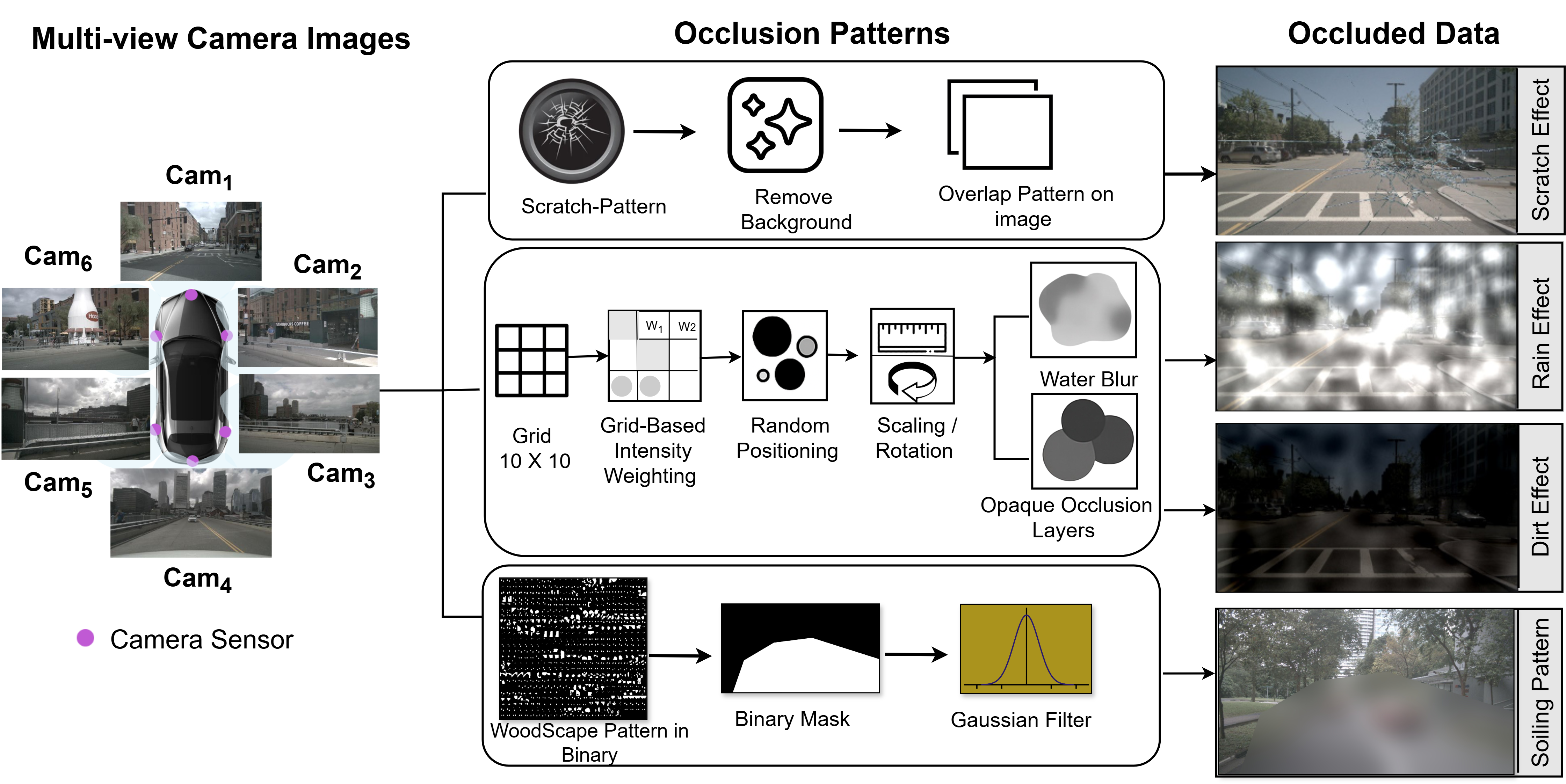}
\caption{\textbf{Overview of the camera occlusion generation pipeline. Multi-view images from the nuScenes dataset \cite{caesar2020nuscenes} are processed to simulate four types of occlusion: (1) scratch effects created \cite{pngegg_scratch} using overlay masks with background removal, (2) rain effects generated \cite{10900335} through grid-based occlusion maps with randomised patch placement, scaling, and rotation, (3) dirt effects introduced \cite{10900335} using opacity-weighted masks on high-impact regions, and (4) soiling patterns derived from WoodScape \cite{yogamani2019woodscape} binary masks smoothed with a Gaussian filter. These occlusions mimic realistic visibility challenges for evaluating the robustness of perception.}} 
\label{fig1}
\end{figure*}

This degradation simulates the accumulation of dirt or dust on the camera lens, as shown in Fig.~\ref{fig1}. In real automotive cameras, dirt commonly builds up on the lens surface due to mud splashes, dust, or road debris during driving, gradually obscuring parts of the field of view.
The image is divided into a $10 \times 10$ grid, and within each cell, occlusion patches are randomly projected. 
To reflect how dirt becomes more prominent in bright areas, patch intensity is weighted according to the local brightness of the underlying region.  

Three obstruction layers, denoted as $M_k$ ($k = 0, 1, 2$), are applied to represent varying occlusion strengths. 
Each layer is randomly scaled, rotated, and positioned across the image to introduce spatial variability. 
The final occluded image is computed as:  
\[
I'(x, y) = \min\Big(\max\big(I(x, y) + \alpha \cdot \sum_{k=0}^2 M_k(x,y),\ 0\big),\ 255\Big),
\]
where $I(x,y)$ is the original pixel value, $\alpha \in [0,1]$ is a global opacity parameter (implemented as \texttt{OPACITY} in code), 
and the min–max operation (\texttt{clip}) ensures that pixel intensities remain within the valid range $[0,255]$.  

By varying $\alpha$ between 0.1 and 0.3 and adjusting patch density, we generate three severity levels corresponding to light, moderate, and heavy dirt conditions. 
This formulation provides a controlled and reproducible framework for assessing the robustness of camera-based perception models under progressive lens contamination.

\subsubsection{Water-Blur Effect Simulation}

This degradation simulates the visual distortions caused by water droplets or condensation on the camera lens, as shown in Fig.~\ref{fig1}. In real automotive cameras, such effects occur when raindrops or moisture accumulate on the lens surface, causing light scattering, smearing, and blurring that obscure image details.
The effect is implemented using the BODA obstruction model~\cite{10900335}, which combines convolution-based directional blurring with random droplet overlays to mimic realistic wet-lens artifacts such as smearing and streaking.  

The model first applies a directional blur to the input image using a kernel $K$ derived from a droplet obstruction pattern:  
\[
\tilde{I}_c(x, y) = \sum_{u,v} I_c(x-u, y-v) \cdot K(u,v), \quad c \in \{R, G, B\},
\]
where $I_c(x,y)$ is the original image in channel $c$ and $K(u,v)$ is the normalized blur kernel. 
The kernel is randomly rotated and scaled to introduce variation in streak orientation and spread, thereby simulating the refraction of light through uneven water films.  

Next, an obstruction layer $O(x,y)$, composed of randomly placed and scaled droplet masks, is blended with the blurred image $\tilde{I}(x,y)$ using an opacity parameter $\alpha \in [0,1]$:  
\[
I'(x, y) = (1 - \alpha)\, \tilde{I}(x, y) + \alpha\, O(x, y).
\]
Here, $\alpha$ controls the strength of the occlusion, determining how visible the droplet layer appears on the final image. 
By adjusting $\alpha$ to 0.1, 0.2, and 0.3, we generate three occlusion severity levels corresponding to mild, moderate, and strong degradation. This process reduces image sharpness and local contrast while maintaining global scene structure, producing realistic approximations of camera degradation under wet or foggy conditions.

\begin{figure*}[ht]
\centering
\includegraphics[width=0.8\textwidth]{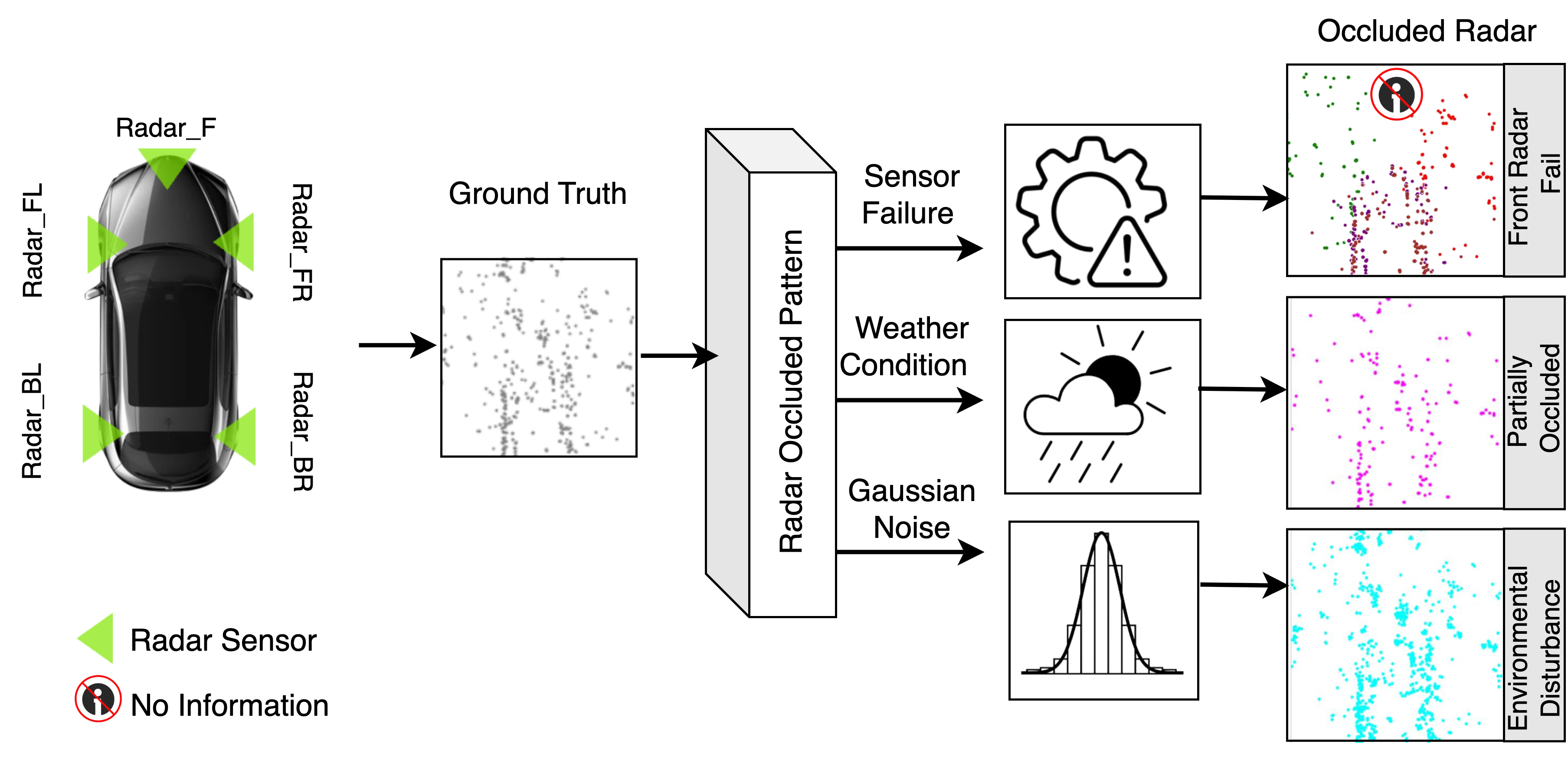}
\caption{\textbf{Overview of the Radar occlusion generation pipeline. Ground truth radar point clouds from five nuScenes sensors \cite{caesar2020nuscenes} are altered using three occlusion types: (1) single sensor failure by removing data from selected radars, (2) weather-induced partial masking of returns, and (3) environmental disturbances via additive Gaussian noise.}} 
\label{fig2}
\end{figure*}

\subsubsection{Scratch Pattern Overlay}

This degradation simulates physical scratches on the camera lens by overlaying realistic scratch textures onto the original images, as shown in Fig.~\ref{fig1}. In real automotive cameras, scratches may result from debris impact during storms or long-term exposure to extreme temperature fluctuations, which can damage the protective lens coating and distort captured visuals. A collection of scratch patterns with varying severities was obtained from publicly available sources. 
Each texture was preprocessed to remove its background, leaving only the visible scratch marks. 
A sample scratch pattern is available here \cite{pngegg_scratch}.  

The occluded image is generated by alpha blending the scratch texture $S$ with the original image $I$:  
\[
I'(x, y) = (1 - \alpha(x,y))\, I(x, y) + \alpha(x,y)\, S(x, y)
\]
where $\alpha(x,y) \in [0,1]$ is the transparency value provided by the scratch texture. 
This blending ensures that all pixel values remain within the valid range $[0,255]$ without the need for clipping.  

This method introduces thin linear artefacts that resemble real lens scratches while keeping the overall scene content visible.

\subsubsection{Woodscape Soiling Patterns Effect}

To emulate lens contamination effects in the nuScenes camera images, we apply synthetic occlusions derived from the WoodScape Soiled dataset \cite{yogamani2019woodscape}. In real automotive cameras, fog, mist, or humidity can form thin translucent layers on the lens \cite{bijelic2020seeing}, reducing contrast and creating hazy, diffused visuals similar to those reproduced by the WoodScape soiling patterns, as shown in Fig.~\ref{fig1}. Each soiling pattern is converted into a binary mask $M \in \{0,1\}^{H \times W}$, where 1 indicates occluded pixels. 
A Gaussian filter $G_{\sigma}$ is then applied to soften the mask boundaries, producing a weighted mask $M' \in [0,1]^{H \times W}$. The blurred region is computed as:  
\[
I_{\text{blurred}} = G_{\sigma} * (I \circ M')
\]
where $I$ is the original image, $G_{\sigma}$ is the Gaussian kernel, $\circ$ denotes element-wise multiplication, and $*$ is the convolution operator.  

The final occluded image is obtained by combining the blurred occluded regions with the original unoccluded content:  
\[
I' = I \circ (1 - M') + I_{\text{blurred}}
\]
This approach introduces realistic occlusion effects while preserving the overall scene structure, enabling the controlled evaluation of perception models under lens-soiling conditions.

\subsection{Radar Noise/Failure Models}
This section describes three types of synthetic occlusions applied to the radar data in the nuScenes dataset. These occlusion strategies simulate real-world challenges such as complete single-sensor failure, partial signal loss, and environmental interference. Each type is designed to test the robustness of radar-based perception under degraded sensing conditions.

\subsubsection{Single-Sensor Failure}

To simulate the failure of an individual radar unit, one of the five nuScenes radar sensors is randomly disabled for each frame, as shown in Fig.~\ref{fig2}. 
The sensor set is defined as:  
\[
\mathcal{R} = \Big\{
\begin{aligned}
&\text{FRONT},\ \text{FRONT\_LEFT},\ \text{FRONT\_RIGHT}, \\
&\text{BACK\_LEFT},\ \text{BACK\_RIGHT}
\end{aligned}
\Big\}
\]
For each sample, a sensor $r_{\text{drop}} \in \mathcal{R}$ is selected at random and excluded:  
\[
\mathcal{R}' = \mathcal{R} \setminus \{r_{\text{drop}}\}, \quad r_{\text{drop}} \sim \mathcal{R}.
\]
Only the remaining sensors in $\mathcal{R}'$ contribute to the point cloud. 
This setup reflects a hardware malfunction or communication dropout of a single radar sensor and enables controlled testing of the perception model's sensitivity to the loss of individual radar views.  

\begin{figure*}[ht]
\centering
\includegraphics[width=0.8\textwidth]{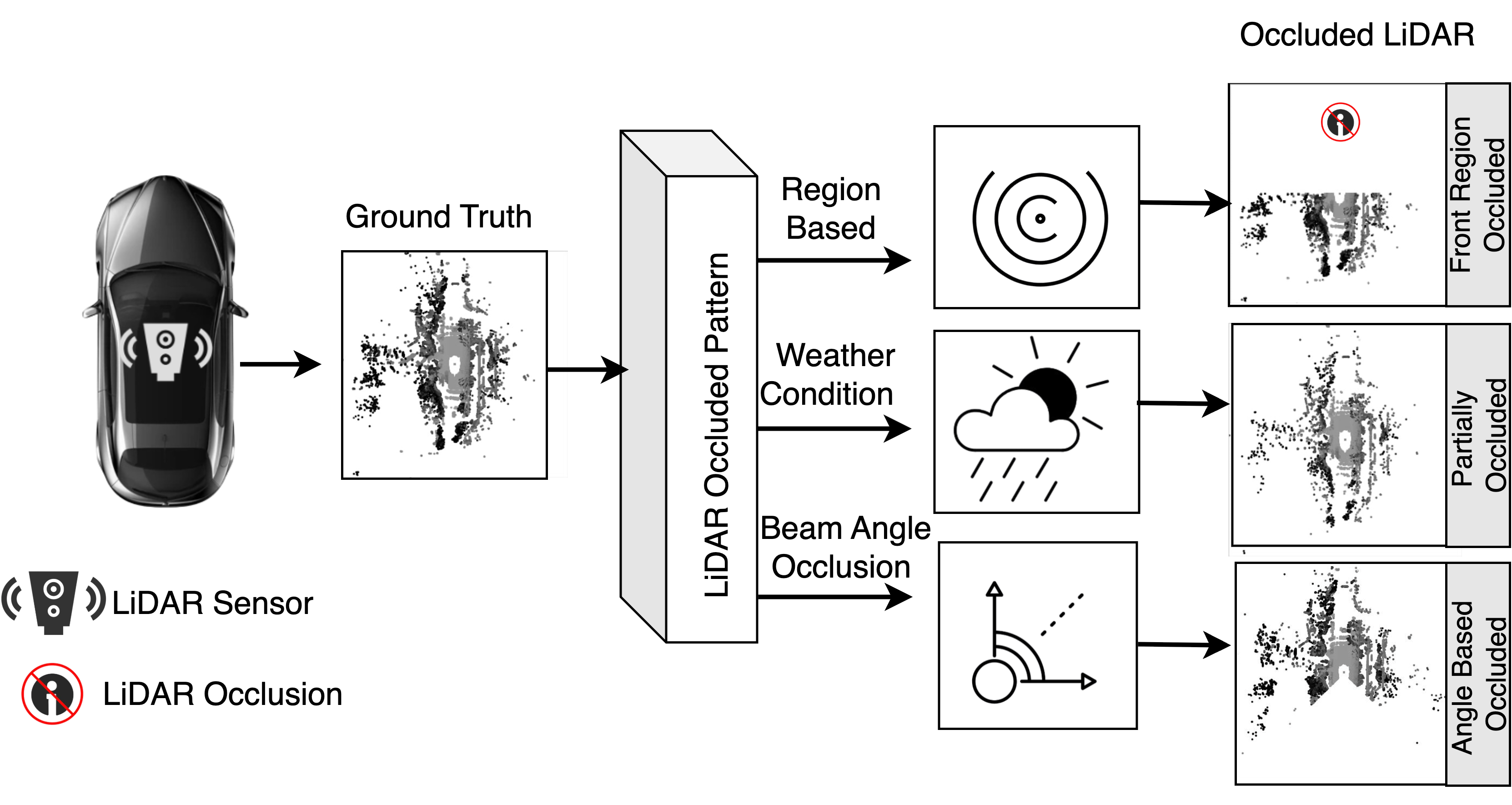}
\caption{\textbf{Overview of the LiDAR occlusion generation pipeline. Ground truth point clouds from the nuScenes dataset \cite{caesar2020nuscenes} are modified using three occlusion types: (1) region-based removal of specific areas, (2) weather-induced partial masking, and (3) beam angle occlusion based on sensor direction.}} 
\label{fig3}
\end{figure*}

\subsubsection{Radar Point Dropout}

This degradation simulates situations where radar measurements are incomplete, such as when signals are blocked by obstacles, affected by reflective interference, or attenuated by environmental clutter, as shown in Fig.~\ref{fig2}. 
To reproduce this effect, a user-defined percentage of radar point returns is randomly discarded from one sensor. 
The dropout is applied uniformly, meaning each point has an equal probability of being removed.  

The severity is governed by a drop percentage parameter $p \in [0, 100]$, which specifies the fraction of points to discard. 
For example, if $p = 30$, then approximately $30\%$ of radar points are removed. 
The number of retained radar points is given by:  
\[
N_{\text{retained}} = N \times \left(1 - \frac{p}{100}\right)
\]
where $N$ is the original number of radar points. 
This parameterised setup provides a controlled way to study perception model performance under varying levels of radar sparsity.  

\subsubsection{Environmental Noise}

This degradation simulates environmental interference by adding Gaussian noise to the radar point cloud, as shown in Fig. \ref{fig2}. 
Such perturbations mimic the effects of heavy rain and fog, which distort radar signals in real-world conditions \cite{zang2019impact}. 
Noise is applied independently to the spatial coordinates $(x, y, z)$ of each radar return, with the perturbation drawn from a zero-mean normal distribution with variance $\sigma^2$.  

Formally, the noisy radar point is given by:  
\[
\mathbf{p}' = \mathbf{p} + \boldsymbol{\epsilon}, 
\quad \boldsymbol{\epsilon} \sim \mathcal{N}(0, \sigma^2 \mathbf{I})
\]
where $\mathbf{p} = (x, y, z)$ is the original point, $\mathbf{p}'$ is the perturbed point, $\sigma$ controls the intensity of the noise and $\mathcal{N}(0, \sigma^2 \mathbf{I})$ denotes a multivariate normal distribution with zero mean and covariance $\sigma^2 \mathbf{I}$. 
This formulation introduces random displacements in 3D space, degrading the geometric precision of radar measurements.

\subsection{LiDAR Occlusion}

To enable the evaluation of LiDAR-based perception robustness, we provide three types of synthetic occlusion applied to the point cloud data. These methods simulate realistic degradations that can arise in automated driving scenarios and target different aspects of LiDAR sensing. The occlusion strategies are: region-based occlusion, partial point cloud dropout, and angle-based occlusion.

Together, these methods provide complementary perspectives on LiDAR degradation, enabling controlled benchmarking of perception models under varied occlusion conditions.

\subsubsection{Region-Based LiDAR Occlusion}

This occlusion simulates directional LiDAR blind spots by removing all points that fall within a specified spatial region relative to the ego vehicle, as shown in Fig. \ref{fig3}. 
The four predefined regions are: front ($x > 0$), back ($x < 0$), left ($y < 0$), and right ($y > 0$).

Formally, a binary mask $M_{\text{region}}(i)$ is defined for each point $i$ as:  
\[
M_{\text{region}}(i) = 
\begin{cases}
1 & \text{if point $i$ lies in the selected region}, \\
0 & \text{otherwise}.
\end{cases}
\]
All points where $M_{\text{region}}(i) = 1$ are removed from the point cloud. 
This setup enables controlled simulation of directional blind spots, mimicking real-world occlusions caused by large vehicles, infrastructure, or other obstacles.  

\subsubsection{Partial LiDAR Point Dropout}

This occlusion simulates the effect of adverse weather conditions, such as fog or heavy rain by randomly discarding a fraction of LiDAR points from each sweep. The dropout is applied uniformly, meaning every point has the same probability of being removed regardless of its spatial location, as shown in Fig. \ref{fig3}. The severity of occlusion is controlled by a parameter $p \in [0, 100]$, which specifies the percentage of points to drop. The number of retained points is then given by:  
\[
N_{\text{retained}} = N \times \left(1 - \frac{p}{100}\right)
\]
Where $N$ is the original number of LiDAR points in the sweep.  
This controlled sparsity enables parameterised evaluation of perception model robustness under degraded LiDAR input.

\subsubsection{Angle-Based LiDAR Occlusion}

This occlusion simulates directional LiDAR blind spots within a conical region relative to the ego vehicle. 
Unlike region-based occlusion, which removes all points in a half-plane (front, back, left, or right). 
The angle-based method allows finer control by excluding points within a user-defined angular range, as shown in Fig. \ref{fig3}. For each LiDAR point, the azimuthal angle is computed as:  
\[
\theta = \tan^{-1}\left(\frac{y}{x}\right)
\]
A point is removed if it lies in the selected spatial region and its azimuth falls within the angular range. 
For example, for a front-facing occlusion with angle $\alpha$, points satisfying are discarded. 
\[
x > 0 \quad \text{and} \quad |\theta| \leq \frac{\alpha}{2}
\]
This produces a conical exclusion zone that mimics real-world occlusions caused by vehicles, infrastructure, or other obstacles in specific directions.

\begin{table}[h!]
\centering
\caption{Summary of occlusion types applied to Camera, Radar, and LiDAR modalities, along with corresponding parameters and their ranges/settings.}
\scalebox{0.80}{
\begin{tabular}{llll}
\toprule
\textbf{Modality} & \textbf{Occlusion Type} & \textbf{Parameters Used} & \textbf{Range / Setting} \\
\midrule
\multirow{2}{*}{Camera} 
    & Dirt and Water-blur      & Opacity                         & 0.1 – 0.3 \\
    & Soiling Mask       & Kernel sizes ($\sigma$)      &  \makecell{(15$\times$15), (51$\times$51),\\(101$\times$101), (251$\times$251)} \\ \\

\multirow{3}{*}{Radar}  
    & Sensor Drop       & Random selection                & Drop 1 of 5 radars \\
    & Point Dropout     & Drop (\%)                       & 0\% – 99\% \\
    & Gaussian Noise    & Std Dev (m)                     & 0.1 – 2 \\\\

\multirow{3}{*}{LiDAR}  
    & Region Drop       & Spatial Region & front/back/left/right \\
    & Angle Occlusion   & Region + Angle (deg)            & eg. Front (30$^\circ$), 60$^\circ$, 90$^\circ$ \\
    & Random Dropout    & Drop (\%)                       & 0\% – 99\% \\
\bottomrule
\end{tabular}
\label{tab:occlusion_params}
}
\end{table}

\section*{VALIDATION AND QUALITY} 

The base dataset used in this work is the nuScenes \cite{caesar2020nuscenes} dataset, which is widely recognised for its high-quality, synchronised multi-modal sensor data and accurate calibration parameters. These properties ensure a reliable foundation for introducing controlled occlusions. To validate the occlusion transformations introduced in this dataset, we carried out a series of quality checks, as discussed below.\\  

First, \textbf{structural consistency} was verified by confirming that all occluded data preserved the original nuScenes format, file integrity, and annotation alignment. Calibration metadata, image resolution, and point cloud geometry remained unchanged to guarantee compatibility with the existing nuScenes dataset.  

\begin{figure}
\centerline{\includegraphics[width=3.5in]{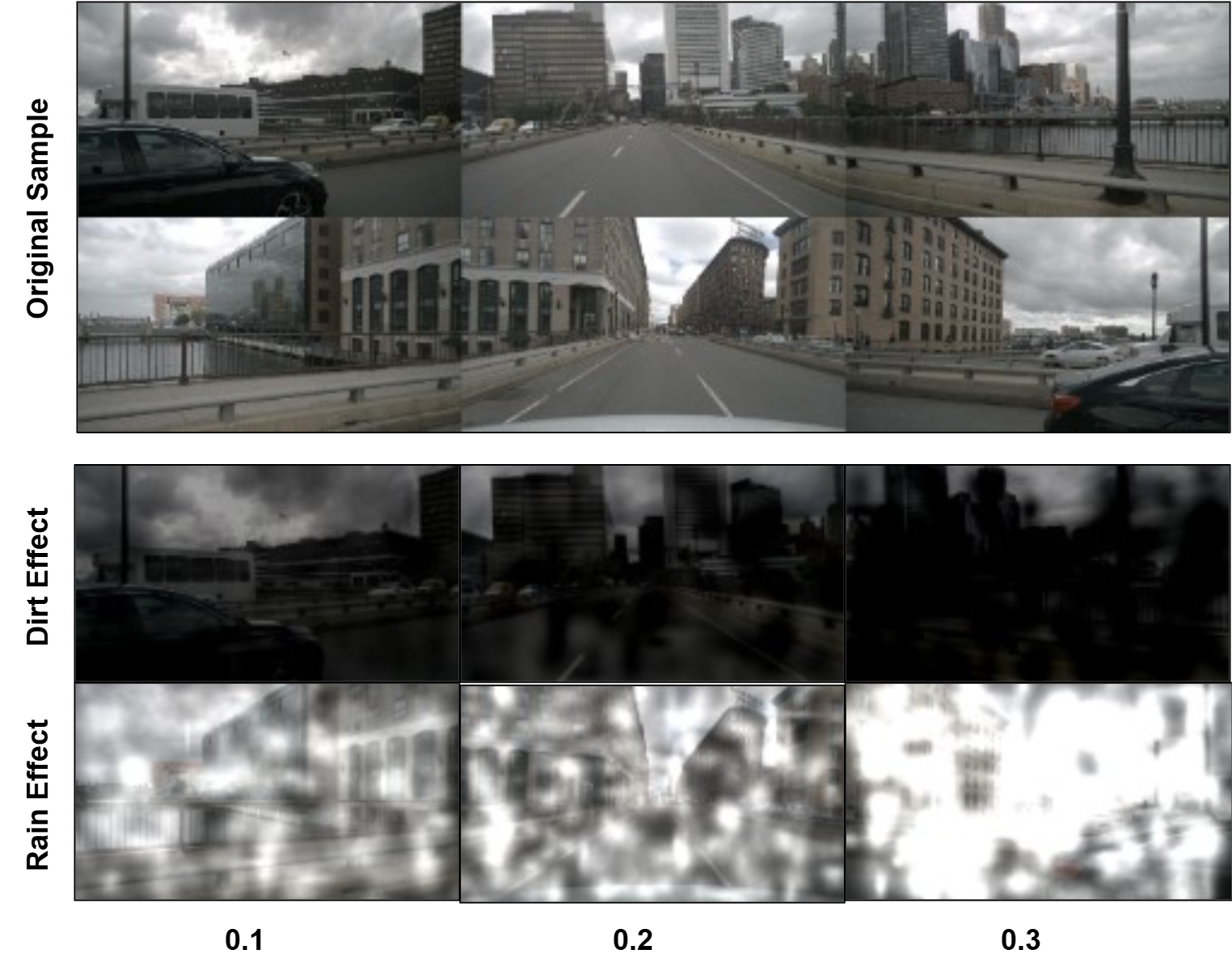}}
\caption{Examples of camera occlusion effects on nuScenes samples. The top row displays the original multi-view images, while the bottom rows illustrate the effects of dirt and rain applied at three increasing occlusion levels (0.1, 0.2, 0.3).\label{fig4}}
\end{figure}

\begin{figure}
\centerline{\includegraphics[width=3.5in]{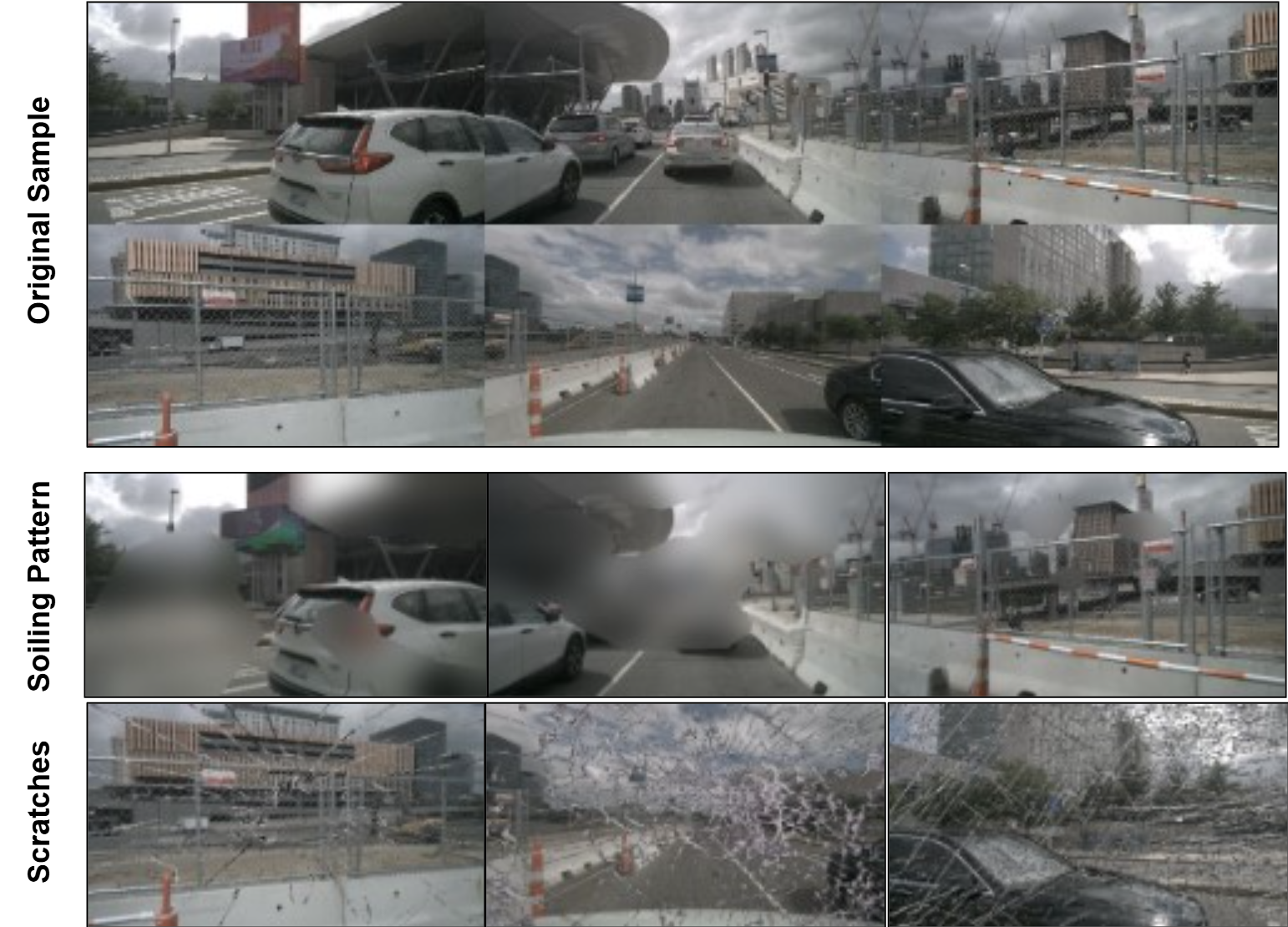}}
\caption{Examples of camera occlusion effects on nuScenes samples. The top row shows the original multi-view images, while the bottom rows illustrate occlusions generated using soiling patterns from WoodScape and scratch overlays, respectively.\label{fig5}}
\end{figure}

\begin{figure}
\centerline{\includegraphics[width=3.5in]{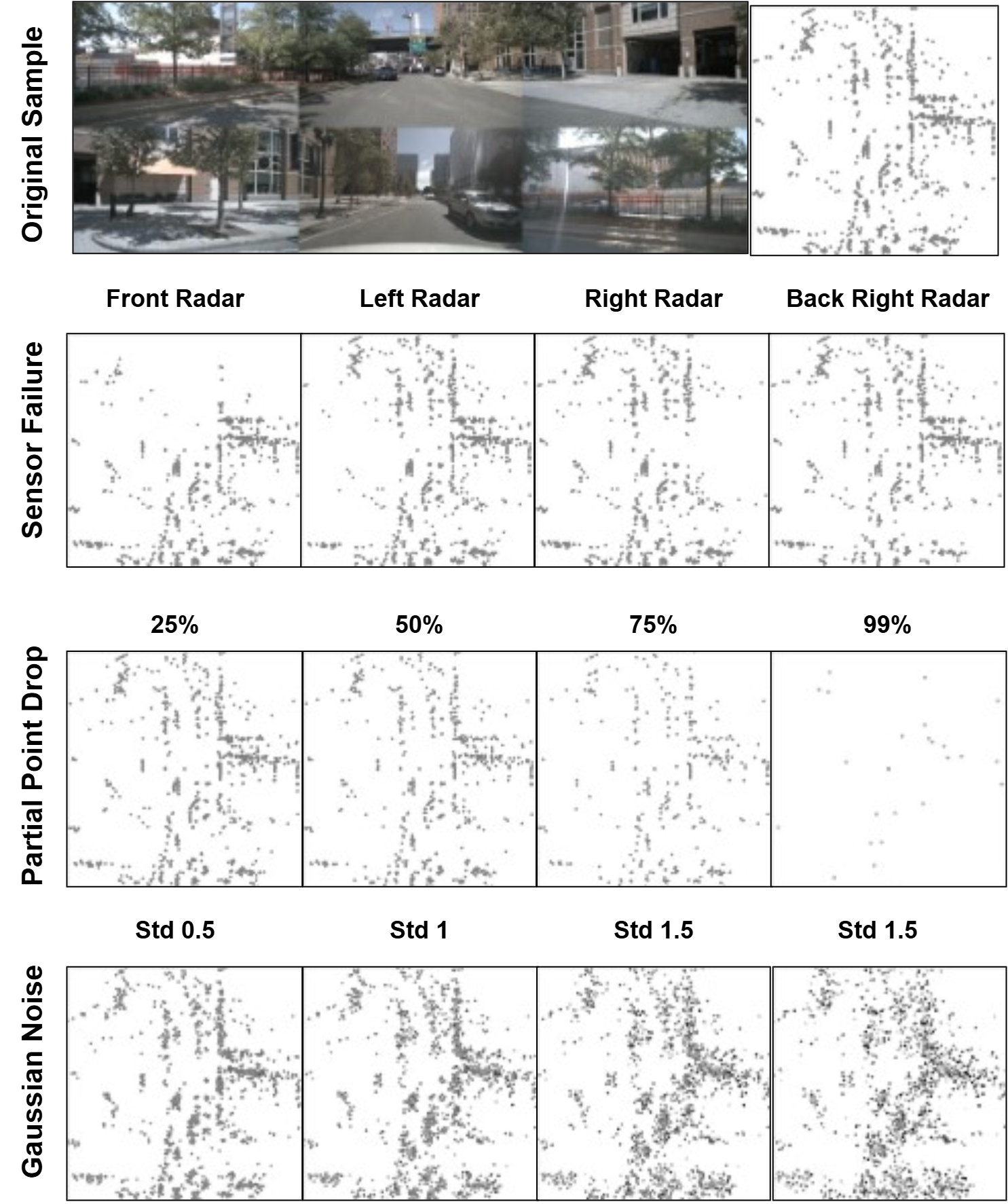}}
\caption{Examples of Radar occlusion effects on nuScenes samples. The top row shows the original sample with radar point clouds, while the following rows illustrate three occlusion types: complete single sensor failure (removal of selected radars), partial point drop at varying levels (25–99\%), and additive Gaussian noise with increasing standard deviation.\label{fig6}}
\end{figure}

\begin{figure}
\centerline{\includegraphics[width=3.5in]{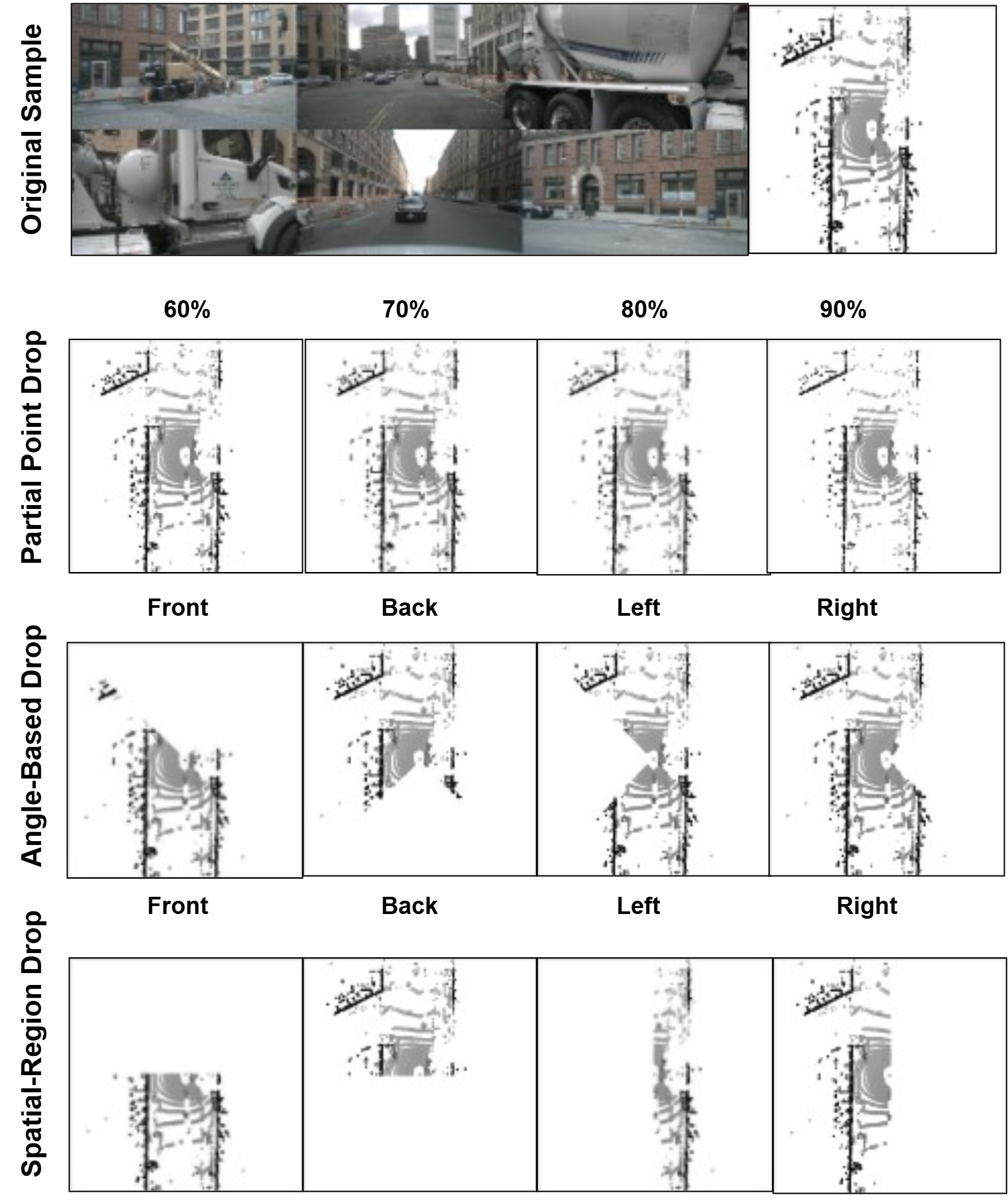}}
\caption{Examples of LiDAR occlusion effects on nuScenes samples. The top row shows the original sample with point clouds, while the following rows illustrate three occlusion types: partial point drop at different levels (60–90\%), angle-based occlusion for front, back, left, and right sectors, and spatial region-based occlusion applied to specific areas.\label{fig7}}
\end{figure}

Second, \textbf{visual inspection} was carried out across all three sensor modalities to verify the realism of the applied occlusions. 
For the camera streams, we reviewed samples with dirt, water-blur, soiling patterns, and scratch overlays at different severity levels, confirming that the transformations produced natural-looking degradations while preserving overall scene structure (Figs. \ref{fig4}, \ref{fig5}). 
For radar, visual checks were performed on examples of complete single sensor failure, partial point dropout, and Gaussian noise injection. As shown in Fig. \ref{fig6}, these effects manifest as missing radar views, reduced point density, or dispersed point distributions, approximating plausible sensor interference in adverse conditions.
For LiDAR, inspection of region-based, partial dropout, and angle-based occlusions confirmed that the expected spatial sectors or angular cones were consistently removed, producing interpretable blind spots in the point cloud (Fig. \ref{fig7}).    

Third, \textbf{parameter verification} was carried out to confirm that the applied occlusions behaved consistently with their defined severity levels across all sensor modalities. 
For camera occlusions, increasing opacity or distortion parameters resulted in visibly stronger degradations, progressing from light to heavy severity. 
For radar and LiDAR occlusions, higher dropout percentages corresponded to proportionally fewer retained points, while larger noise parameters produced increasingly dispersed point distributions. 
Representative examples are shown in Figs. \ref{fig4}, \ref{fig5}, \ref{fig6}, and \ref{fig7}, demonstrating that the parameterised controls produce predictable and reproducible degradation patterns. 
This confirms that the occlusion pipeline reliably translates user-defined parameters into measurable effects across all sensor streams.

Finally, \textbf{practical validation} was performed by evaluating the occluded data on three downstream perception tasks, as shown in Table \ref{tab:impact_analysis}. (i) vehicle segmentation \cite{kumar2025minimizing}, \cite{11204511}, (ii) map segmentation \cite{11204511}, and (iii) 3D object detection \cite{kumar2025Evaluating}. These tasks are fundamental perception problems in automated driving \cite{feng2020deep}, encompassing object-level detection, scene understanding, and spatial reasoning. Baseline architectures, including SimpleBEV \cite{harley2022simple}, BEVfusion \cite{liu2022bevfusion}, and BEVCar \cite{schramm2024bevcar}, were tested. In all cases, measurable degradation in performance was observed, confirming that the occlusions simulate sensor-level disturbances and provide meaningful challenges for robustness benchmarking.

Moreover, to assess their perceptual impact and ensure reproducibility, we computed the \gls{ssim}~\cite{wang2004image} between clean and occluded images using 5k random samples per camera. \gls{ssim} was chosen because it captures perceptual degradations in structural and contrast information, providing a more reliable measure of visual quality loss than pixel-based metrics such as MSE or PSNR. Dirt occlusion produced opaque patches that completely masked pixel regions, resulting in larger structural distortions compared to water-blur, which primarily smeared image details while preserving the overall scene structure. 

The mean \gls{ssim} drops for dirt were 0.43, 0.73, and 0.88, while for water-blur they were 0.28, 0.29, and 0.45 at occlusion levels 0.1, 0.2, and 0.3, respectively. For scratches, the mean drop was 0.34, computed over randomly applied patterns of varying severities. For WoodScape soiling patterns, it was 0.074, obtained by randomly applying kernels of different sizes (15×15, 51×51, 101×101, and 251×251) across images. The relatively small SSIM drop for the WoodScape soiling effect reflects its translucent, low-opacity nature, which simulates subtle lens haze rather than strong occlusion.

Together, these checks establish the validity and quality of the occluded nuScenes dataset, ensuring it is both reliable for research and impactful for evaluating perception robustness in automated driving.  

\begin{table}[h]
\centering
\caption{Performance degradation of perception models under occluded data.}
\scalebox{0.88}{
\begin{tabular}{l l c c c}
\toprule
\textbf{Model} & \textbf{Task (Metric)} & \textbf{Clean} & \textbf{With Occlusion} & \textbf{Drop (\%)} \\
\midrule
SimpleBEV   & Vehicle Seg (IoU)   & 47.4 & 34.3 & -27.6 \\
BEVCar      & Map Seg (mIoU)      & 70.9 & 60.1 & -15.2 \\
BEVFusion   & Object Det (mAP)       & 56.6 & 48.6 & -14.1 \\
\bottomrule
\end{tabular}
}
\label{tab:impact_analysis}
\end{table}

\section*{RECORDS AND STORAGE}

The Occluded nuScenes dataset is structured to mirror the original nuScenes \cite{caesar2020nuscenes} format, ensuring compatibility with existing perception pipelines and annotation files. 
All occluded data follows the original nuScenes naming conventions and timestamp alignment, allowing users to substitute occluded samples in place of the originals for evaluation without additional preprocessing.  

\textbf{Camera records} 
For both the full and mini nuScenes versions, occluded images have been generated for all six camera views and stored as JPEG files. 
Each occlusion type (dirt, water-blur, soiling patterns, scratches) is provided at multiple severity levels. 
The folder structure mirrors the original nuScenes format (e.g., \texttt{samples/CAM\_FRONT/}), enabling direct plug-and-play with existing pipelines.  

\textbf{Radar and LiDAR records} 
Due to the large storage overhead of generating occluded point cloud files across all sweeps, pre-computed radar and LiDAR occlusion data are not distributed. 
Instead, we provide parameterised scripts that allow users to apply occlusions on demand. 
Radar occlusions include single sensor failure, partial point dropout, and Gaussian noise, while LiDAR occlusions include region-based, partial dropout, and angle-based methods. 
This design choice ensures reproducibility and flexibility without requiring excessive storage space.  

\textbf{Annotations} 
All original nuScenes annotations, including 3D bounding boxes, object categories, tracking information, and map layers, are preserved without modification. This choice guarantees full compatibility with existing nuScenes evaluation protocols.  

\textbf{Storage and Access} 
The occluded camera dataset requires a total of 310.31 GB for the full version and 3.95 GB for the mini version. These figures correspond to the combined size of all four occlusion types applied at multiple severity levels. The storage size for each individual occlusion type is therefore smaller than the reported totals. 
For radar and LiDAR, we provide parameterised occlusion generation scripts, which are distributed via GitHub together with documentation and usage examples.

\section*{INSIGHTS AND NOTES}

The Occluded nuScenes Dataset provides the first resource that applies controlled and reproducible occlusions across camera, radar, and LiDAR modalities in automated driving.
This resource opens new opportunities for studying perception robustness, benchmarking sensor fusion architectures, and developing methods that can tolerate sensor degradation in realistic conditions.  

The dataset can be used to evaluate the resilience of multi-sensor fusion models, to train and test perception systems under occluded conditions, and to explore strategies for sensor redundancy and fault tolerance. 
It serves as a controlled evaluation resource for safety-critical scenarios, enabling reproducible testing of perception robustness across different model architectures. Only the camera modality is released as pre-generated occluded data, while radar and LiDAR occlusions are provided through scripts to avoid excessive storage requirements. 
As the occlusions are synthetic, they approximate but do not fully replicate real-world sensor degradations. 
Some extreme parameter settings, such as very high dropout rates, are intended as stress tests rather than typical driving scenarios.  

Overall, this dataset supports both benchmarking studies and occlusion-aware training, while also offering the flexibility to generate occlusions at different severity levels for specific applications, all within a reproducible framework.

\section*{SOURCE CODE AND SCRIPTS} 

Two camera occlusion methods (dirt and water-blur) were adapted from a publicly available repository \cite{10900335}, while scratch overlays, WoodScape soiling, and all radar and LiDAR occlusion methods were implemented by the main author. 
All radar and LiDAR scripts are available at \url{https://github.com/D2ICE-Automotive-Research/nuScenes-Camera-Radar-LiDAR-Occlusion}.

\section*{ACKNOWLEDGEMENTS AND INTERESTS}

This dataset is based on the nuScenes dataset by Motional. We thank the creators of nuScenes for providing the original dataset, which we extend by introducing synthetic occlusions across the sensor data.
S.K. created the dataset. C.E. acquired the necessary funding. C.E., E.M.G., and V.D. conceived and supervised the project. S.K. wrote the initial draft of the paper, and all the authors contributed to revisions.  
The authors have declared no conflicts of interest.

\bibliography{reference}

\bibliographystyle{IEEEtran}

\end{document}